\documentclass{article}
\usepackage{spconf,amsmath,graphicx}
\usepackage{amsmath,graphicx}
\usepackage{amsfonts}
\usepackage{color}
\usepackage{caption}
\usepackage{subfigure}

\include{definitions}

\usepackage{amsmath,graphicx}
\usepackage{amsthm}
\usepackage{amssymb}
\usepackage{epstopdf}
\usepackage{subfigure}
\usepackage{color}
\usepackage{caption}
\usepackage{mathtools}
\usepackage{tabulary}
\usepackage{booktabs}
\usepackage{xr}

\usepackage{zref-base}
\usepackage{atveryend}

\newtheorem{theorem}{Theorem}[section]

\newtheorem{lemma}[theorem]{Lemma}
\newtheorem*{lemma*}{Lemma}

\title{Clustering of Data with Missing Entries}
\name{Sunrita Poddar, Mathews Jacob\thanks{This  work  is  supported  by NIH 1R01EB019961-01A1 and ONR-N000141310202.}}
\address{Department of Electrical and Computer Engineering, University of Iowa, IA, USA}

\begin{document}

\maketitle

\begin{abstract}

The analysis of large datasets is often complicated by the presence of missing entries, mainly because most of the current machine learning algorithms are designed to work with full data. The main focus of this work is to introduce a clustering algorithm, that will provide good clustering even in the presence of missing data. The proposed technique solves an $\ell_0$ fusion penalty based optimization problem to recover the clusters. We theoretically analyze the conditions needed for the successful recovery of the clusters. We also propose an algorithm to solve a relaxation of this problem using saturating non-convex fusion penalties. The method is demonstrated on simulated and real datasets, and is observed to perform well in the presence of large fractions of missing entries.

\end{abstract}


\vspace{-1em}

\section{Introduction}
\label{sec:introduction}

Clustering is a popular unsupervised data analysis technique for finding natural groupings in the absence of training data. Specifically, it assigns each data point to a group, such that all points within a group are similar and points in different groups are dissimilar in some sense. Clustering methods are widely used in the analysis of gene expression data, image segmentation, identification of lexemes in handwritten text, search result grouping and recommender systems \cite{saxena2017review,jain1999data}. 

Most clustering algorithms cannot be directly applied to datasets with missing entries. For example, gene expression data often contains missing entries due to image corruption, fabrication errors or contaminants \cite{de2015impact}, rendering gene cluster analysis difficult. Likewise, large databases used by recommender systems (e.g Netflix) usually have a huge amount of missing data, which makes pattern discovery challenging \cite{bell2008bellkor}. Similar issues are reported in the context of missing responses in surveys \cite{brick1996handling} and failing imaging sensors in astronomy \cite{wagstaff2005making} are reported to make the analysis in these applications challenging. The most obvious way to apply existing clustering algorithms to data with missing entries is to convert the data to a complete one. This can be done using deletion or imputation \cite{dixon1979pattern}. An extension of the weighted sum-of-norms algorithm \cite{hocking2011clusterpath} has been proposed where the weights are estimated from the data points by using some imputation techniques on the missing entries \cite{chen2015convex}. A majorize minimize algorithm was introduced to solve for the cluster-centres and cluster memberships in \cite{chi2016k}, which offers proven reduction in cost with iteration. However, these is no theoretical analysis of these algorithms, which makes it difficult to determine what fraction of entries need to be sampled to recover the correct clusters.

In this paper, we introduce an algorithm to cluster data when some of the features are missing in each point. The method is inspired by the recently proposed sum-of-norms clustering technique \cite{hocking2011clusterpath}. This technique assigns a surrogate variable to each data point, which is an estimate of the cluster centre to which that point belongs. When a fusion penalty is used, it is observed that the surrogate variables belonging to the same cluster coalesce to that centre point. These values denote the estimated cluster centres. In prior work, we used a weighted convex fusion penalty to recover under-sampled MRI images lying on a manifold \cite{storm,poddar2014joint}, where the weights were estimated using a special navigator acquisition. In this work, we propose an optimization problem with an $\ell_0$ norm based fusion penalty, since we have observed that non-convex fusion penalties provide better clustering performance than convex ones. The main focus is to theoretically analyze the conditions for the successful recovery of the clusters from data using the proposed optimization technique, when several features are missing. This analysis reveals that the clustering performance is determined by factors such as cluster-separation, cluster variance and feature coherence. When two clusters are distinguishable by only very few features, then it is difficult to distinguish between them if these features are not observed, making feature coherence important. As expected, we also obtain a higher probability of successful clustering in the presence of fewer missing entries. We propose an algorithm to efficiently solve a relaxation of this optimization problem, using saturating non-convex fusion penalties. It is demonstrated on simulated and real datasets that the proposed algorithm performs successful clustering in the presence of large fractions of missing entries.

\vspace{-0.5em}

\section{Clustering using $\ell_0$ fusion penalty}

\subsection{Background}
We consider the clustering of points drawn from one of $K$ distinct clusters $C_1, C_2, \ldots, C_K$. We denote the center of the clusters by $\mathbf c_1, \mathbf c_2, \ldots, \mathbf c_K \in \mathbb R^P$. For simplicity, we assume that there are $M$ points in each of the clusters. The individual points in the $k^{\rm th}$ cluster are modelled as:
\begin{equation}
\label{noisemodel}
\mathbf z_k(m) = \mathbf c_k + \mathbf n_{k}(m); ~~m=1,..,M, ~k=1,\ldots,K
\end{equation}
Here, $\mathbf n_{k}(m)$ is the noise or the variation of $\mathbf z_k(m)$ from the cluster center $\mathbf c_k$. The set of input points $\{\mathbf x_i\},i=1,..,KM$ is obtained as a random permutation of the points $\{\mathbf z_k(m)\}$. The objective of a clustering algorithm is to estimate the cluster labels, denoted by $\mathcal C(\mathbf x_i)$ for $i = 1,..,KM$. 

The sum-of-norms (SON) method is a recently proposed convex clustering algorithm \cite{hocking2011clusterpath}. Here, a surrogate variable $\mathbf u_i$ is introduced for each point $\mathbf x_i$, which is an estimate of the centre of the cluster to which $\mathbf x_i$ belongs. In order to find the optimal $\{\mathbf u_i^*\}$, the following optimization problem is solved:

\begin{equation}
\label{SON}
\{\mathbf u_i^*\} = \arg \min_{\{\mathbf u_i\}} \sum_{i=1}^{KM}\|\mathbf x_i - \mathbf u_i\|_2^2 + \lambda \sum_{i=1}^{KM} \sum_{j=1}^{KM} \|\mathbf u_i - \mathbf u_j\|_{p}
\end{equation}

The fusion penalty ($\|\mathbf u_i - \mathbf u_j\|_{p}$) can be enforced using different $\ell_p$ norms, out of which the $\ell_1$, $\ell_2$ and $\ell_\infty$ norms have been used in literature \cite{hocking2011clusterpath}. The use of sparsity promoting fusion penalties encourages sparse differences $\mathbf u_i-\mathbf u_j$, which facilitates the clustering of the points $\{\mathbf u_i\}$.

\begin{figure}[!t]
	\centering
	\center{\includegraphics[width=0.5\textwidth]{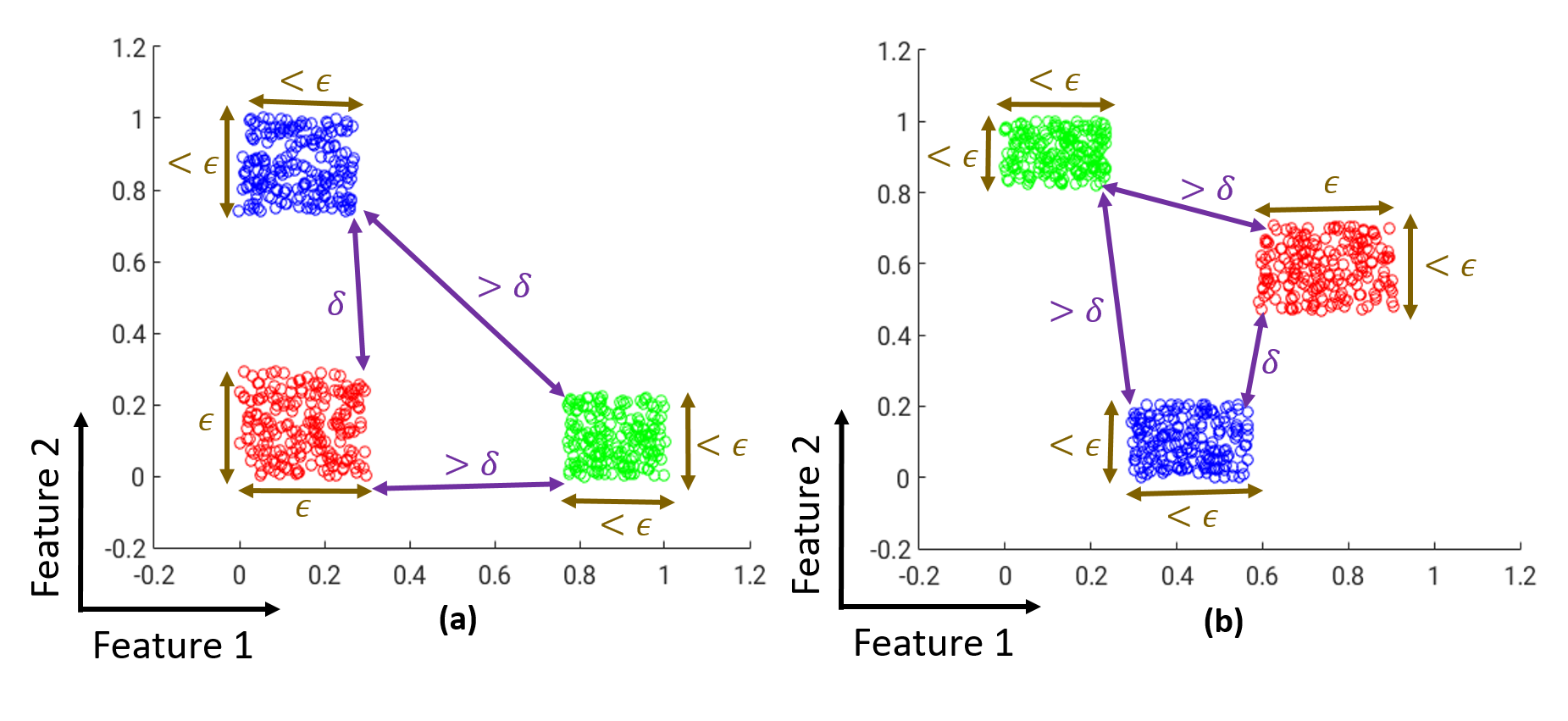}}
	\caption{Central Assumptions: (a) and (b) show different datasets of points $\in \mathbb{R}^2$ lying in 3 clusters (denoted by red, green and blue). A.1 and A.2 are illustrated in both (a) and (b). The importance of A.3 can be appreciated by comparing (a) and (b). In (a), points in the red and blue clusters cannot be distinguished using only feature 1, while the red and green clusters cannot be distinguished using only feature 2. Due to low coherence in (b), this problem does not arise.}
	\label{FigAssump}
\end{figure}

\subsection{Central Assumptions}
We make the following assumptions (illustrated in Fig \ref{FigAssump}), which are key to the successful clustering of the points: 
\begin{description}
\item [A.1:] \textbf{Cluster separation:} Points from different clusters are separated by $\delta >0$ in the $\ell_2$ sense, i.e:
\begin{equation}
\label{deltadef}
\min_{\{m,n\}}\|\mathbf z_{k}(m) -\mathbf z_{l}(n)\|_{2} \geq \delta; ~\forall\; k\neq l
\end{equation}
\item [A.2:] \textbf{Cluster size:} The maximum separation of points within any cluster in the $\ell_{\infty}$ sense is $\epsilon \geq 0$, i.e:
\begin{equation}
\label{epsdef}
\max_{\{m,n\}}\|\mathbf z_{k}(m) - \mathbf z_{k}(n)\|_{\infty} = \epsilon; ~\forall k=1,\ldots,K
\end{equation}
\item [A.3:] \textbf{Feature concentration:} The coherence of a vector $\mathbf y \in \mathbb R^P$ is defined as: $\mu(\mathbf y) = \frac{P\|\mathbf y\|_{\infty}^2}{\|\mathbf y\|_2^2}$. We bound the coherence of the difference between points from different clusters as:
\begin{equation}
\label{coherence}
\max_{\{m,n\}}\mu(\mathbf z_k(m) - \mathbf z_l(n)) \leq \mu_0; ~\forall\; k\neq l
\end{equation}
\end{description}
The quantity $\kappa = \frac{\epsilon\sqrt P}{\delta}$ is a measure of the difficulty of the clustering problem. The recovery of clusters when $\kappa$ is small is expected to be easier. 

\subsection{Theoretical Guarantees}
\label{generalcase}

We study the problem of clustering $\{\mathbf x_i\}$ in the presence of entries missing uniformly at random. We arrange the points $\{\mathbf x_i\}$ as columns of a matrix $\mathbf X$. We assume that each entry of $\mathbf X$ is observed with probability $p_0$. The entries measured in the $i^{th}$ column are denoted by:
\begin{equation}
\mathbf y_i = \mathbf S_i\, \mathbf x_i, ~~ i=1,..,KM
\end{equation}
where $\mathbf S_i$ is the sampling matrix, formed by selecting rows of the identity matrix. 
We consider solving the following optimization problem to obtain the cluster memberships from data with missing entries:
\begin{equation}
\label{l0prob}
\begin{split}
\{\mathbf u_i^{*}\} = & \min_{\{\mathbf u_i\}} \sum_{i=1}^{KM}\sum_{j=1}^{KM}\|\mathbf u_i - \mathbf u_j\|_{2,0}\\ & \mbox{ s.t } \|\mathbf S_i\;(\mathbf x_i - \mathbf u_i)\|_\infty \leq {\frac{\epsilon}{2}}, i\in\{1 \ldots KM\}
\end{split}
\end{equation}
We claim that the above algorithm can successfully recover the clusters with high probability when the clusters are well separated (low $\kappa$), the sampling probability $p_0$ is sufficiently high and the coherence $\mu_0$ is small. We state our theoretical guarantees after defining the following quantities:

\begin{itemize}
\item Upper bound for probability that two points have $< \frac{p_0^2P}{2}$ commonly observed locations: $\gamma_0 \coloneqq (\frac{e}{2})^{-\frac{p_0^2P}{2}}$

\item Given that two points from different clusters have $> \frac{p_0^2P}{2}$ commonly observed locations, upper bound for probability that they can yield the same $\mathbf u$ without violating the constraints in \eqref{l0prob}: $\delta_0 \coloneqq e^{-\frac{p_0^2P(1-\kappa^2)^2}{\mu_0^2}}$

\item Upper bound for probability that two points from different clusters can yield the same $\mathbf u$ without violating the constraints in \eqref{l0prob}: $\beta_0 \coloneqq 1-(1-\delta_0)(1-\gamma_0)$

\item  Upper bound for failure probability of \eqref{l0prob}: $\eta_0 \coloneqq \sum_{\{m_j\} \in \mathcal S}\left[\beta_0^{\frac{1}{2}(M^2-\sum_j {m_j^2})} \prod_j {M \choose m_j}\right]$ where $\mathcal S$ is the set of all sets of positive integers $\{m_j\}$ such that: $2 \leq \mathcal U(\{m_j\}) \leq K$ and $\sum_j m_j = M$. Here, the function $\mathcal U$ counts the number of non-zero elements in a set. For example, if $K=2$ then $\eta_0 =  \sum_{i=1}^{M-1}\left[\beta_0^{i(M-i)} {M \choose i}^2\right]$.
\item For $K=2$ and $\log \beta_0 \leq \frac{1}{M-1} + \frac{2}{M-2}\log \frac{1}{M-1}$, we have $\eta_0 \leq M^3 \beta_0^{M-1} \coloneqq \eta_{0,{\rm approx}}$.

\end{itemize}

\begin{lemma}
\label{cent}
Consider any two points $\mathbf x_1 $ and $\mathbf x_2 $ from the same cluster. A solution $\mathbf u$ exists for the following equations:
	\begin{eqnarray}
\label{samecluster}
\|\mathbf S_i\,(\mathbf x_i-\mathbf u)\|_{\infty} &\leq& {\frac{\epsilon}{2}}; ~ ~i=1,2
\end{eqnarray}
with probability $1$.
\end{lemma}
\begin{lemma}
\label{partDist}
Consider any two points $\mathbf x_1$ and $\mathbf x_2$ from different clusters, and assume that $\kappa<1$. A solution $\mathbf u$ exists for the following equations:
	\begin{eqnarray}
\|\mathbf S_i\,(\mathbf x_i-\mathbf u)\|_{\infty} &\leq& {\frac{\epsilon}{2}}; ~ ~i=1,2
\end{eqnarray}
with probability less than $\beta_0$.
\end{lemma} 
The above lemmas indicate that two points from the same cluster can always be assigned the same centre $\mathbf u^*$. However, for a pair of points from different clusters, this can happen with a probability $<\beta_0$. We note that $\beta_0$ decreases with a decrease in $\kappa$. Using lemmas \ref{cent} and \ref{partDist}, we get the following result for a large number of points from multiple clusters:

\begin{lemma}
\label{smallClusters}
Assume that $\{\mathbf x_i: i\in \mathcal I, |\mathcal I|= M\}$ is a set of points chosen randomly from multiple clusters (not all are from the same cluster). If  $\kappa<1$, a solution $\mathbf u$ does not exist for the following equations:
\begin{equation}
\|\mathbf S_i\,(\mathbf x_i-\mathbf u)\|_{\infty} \leq {\frac{\epsilon}{2}}; ~ ~\forall i \in \mathcal I
\end{equation} 
with probability exceeding $1 - \eta_0$.
\end{lemma}

We note here, that for a low value of $\beta_0$ and a high value of $M$, we will arrive at a very low value of $\eta_0$. Lemma \ref{smallClusters} can be used to arrive at our main result: 

\begin{theorem}
\label{mainResult}
If $\kappa<1$, the solution to the optimization problem \eqref{l0prob} is identical to the ground-truth clustering with probability exceeding $1 - \eta_0$.
\end{theorem}

The reasoning follows from the fact that all solutions with cluster sizes smaller than $M$ are associated with a higher cost than the ground-truth solution. In the special case where there are no missing entries, the constraints of optimization problem \eqref{l0prob} reduce to: $\|\mathbf x_i - \mathbf u_i\|_{\infty} \leq \frac{\epsilon}{2}$. We have the following theorem guaranteeing successful recovery for the clusters:

\begin{theorem}
	\label{noMissingFinal}
	If $\kappa<1$, the solution to the optimization problem \eqref{l0prob} is identical to the ground-truth clustering in the absence of missing entries.
\end{theorem}

\section{Relaxation of the $\ell_0$ penalty}

We propose to solve the following relaxation of the optimization problem \eqref{l0prob}, which is more computationally feasible:
\begin{equation}
\label{relax1}
\begin{split}
\{\mathbf u_i^*\} & = \arg \min_{\{\mathbf u_i\}} \sum_{i=1}^{KM}\|\mathbf S_i(\mathbf u_i - \mathbf x_i)\|_2^2 \\ & + \lambda \sum_{i=1}^{KM} \sum_{j=1}^{KM}\phi(\|\mathbf u_i - \mathbf u_j\|_2)
\end{split}
\end{equation}
Here $\phi$ is a function approximating the $\ell_{0}$ norm, such as:
\begin{itemize}
	\item $\ell_p$ norm: $\phi(x) = |x|^p$, for some $0<p<1$.
	\item $H_1$ penalty: $\phi(x) = 1-e^{-\frac{x^2}{2\sigma^2}}$.
\end{itemize}
Similar to \cite{mohsin2015iterative, wendy}, we reformulate the problem by majorizing the penalty $\phi$ using a quadratic surrogate functional: $\phi(x) \leq w(x) x^2 + d$, where $w(x) = \frac{\phi^{'}(x)}{2x}$, and $d$ is a constant. We now state the majorize-minimize formulation for problem \eqref{relax1} as:
\begin{equation}
\begin{split}
\label{relaxConstrained}
\{\mathbf u_i^*, w_{ij}^*\} & = \arg \min_{\{\mathbf u_i,w_{ij}\}}\sum_{i=1}^{KM}\| \mathbf S_i(\mathbf u_i - \mathbf x_i)\|_2^2\\ + & \lambda \sum_{i=1}^{KM}\sum_{j=1}^{KM} w_{ij}\|\mathbf u_i - \mathbf u_j\|_2^2
\end{split}
\end{equation}
We solve problem \eqref{relaxConstrained} by alternating between minimization with respect to $\{\mathbf u_i\}$ and $\{w_{ij}\}$ till convergence. 

\vspace{-1em}

\section{Results}

\subsection{Study of Theoretical Guarantees}

\begin{figure}
	\centering
	\center{\includegraphics[width=0.48\textwidth]{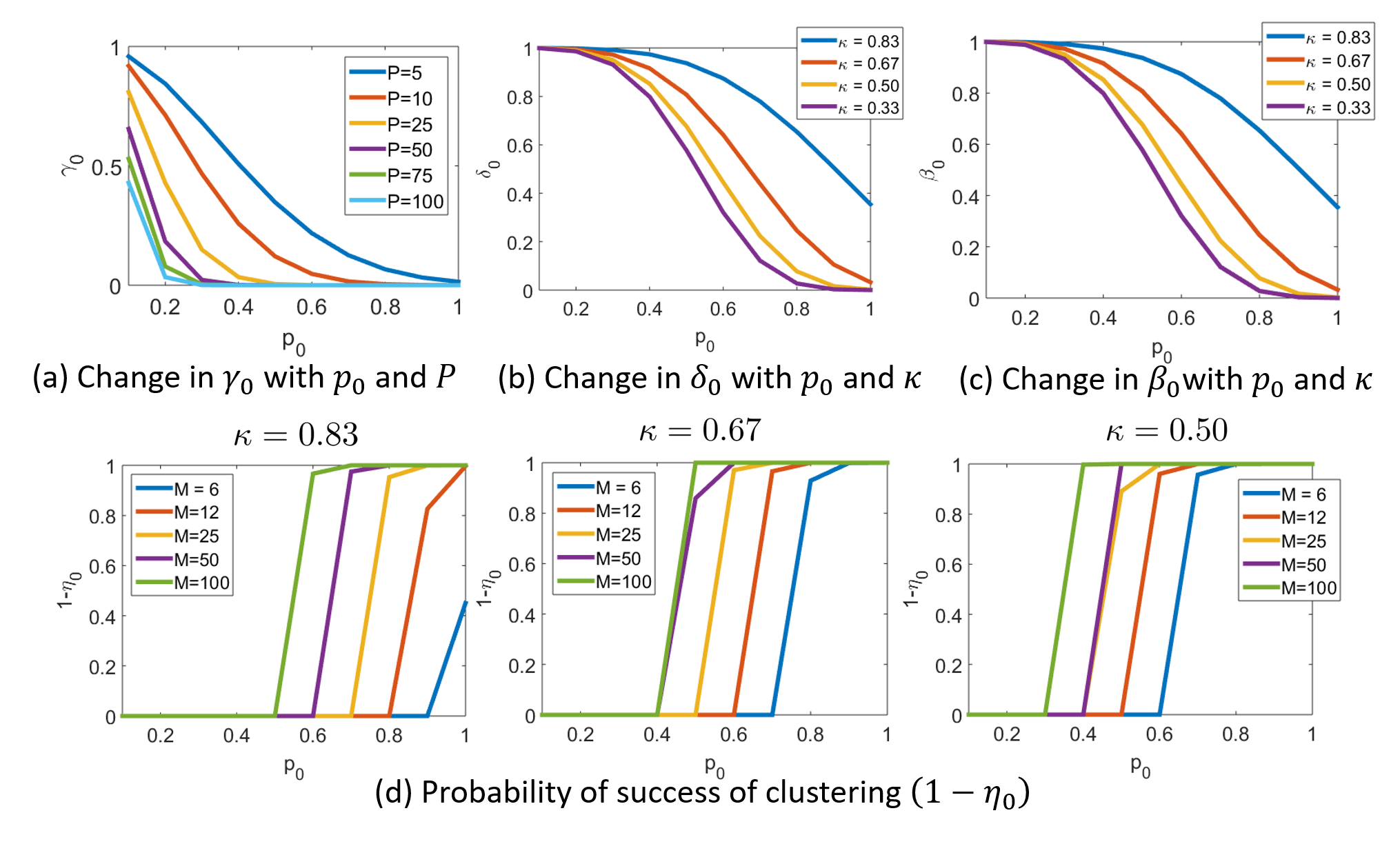}}
	\caption{Study of Theoretical Guarantees. Quantities $\gamma_0, \delta_0$ and $\beta_0$ defined in Section \ref{generalcase} are studied in (a), (b) and (c). In (b), (c) and (d), $P=50$ and $\mu_0=1.5$. As expected, $\beta_0$ decreases with increase in $p_0$ and decrease in $\kappa$. Considering $K=2$ clusters, a lower bound for the  probability of successful clustering $(1-\eta_0)$ is shown in (d) for different $\kappa$.}
	\label{FigTh}
\end{figure}

We observe the behaviour of $\gamma_0, \delta_0, \beta_0$ and $\eta_0$ as a function of $p_0, P, \kappa$ and $M$. In Fig \ref{FigTh} (a), the change in $\gamma_0$ is shown as a function of $p_0$ for different values of $P$. In subsequent plots, we fix $P=50$ and $\mu_0=1.5$. In Fig \ref{FigTh} (b), the change in $\delta_0$ is shown as a function of $p_0$ for different values of $\kappa$. In Fig \ref{FigTh} (c), the behaviour of $\beta_0$ is shown. We consider $K=2$ for subsequent plots. $(1-\eta_0)$ is plotted in (d) as a function of $p_0$ for different values of $\kappa$ and $M$. As expected, the probability of success of the clustering algorithm increases with decrease in $\kappa$ and increase in $p_0$ and $M$.

\subsection{Clustering of Simulated Data}

\begin{figure}
	\centering
	\center{\includegraphics[width=0.48\textwidth]{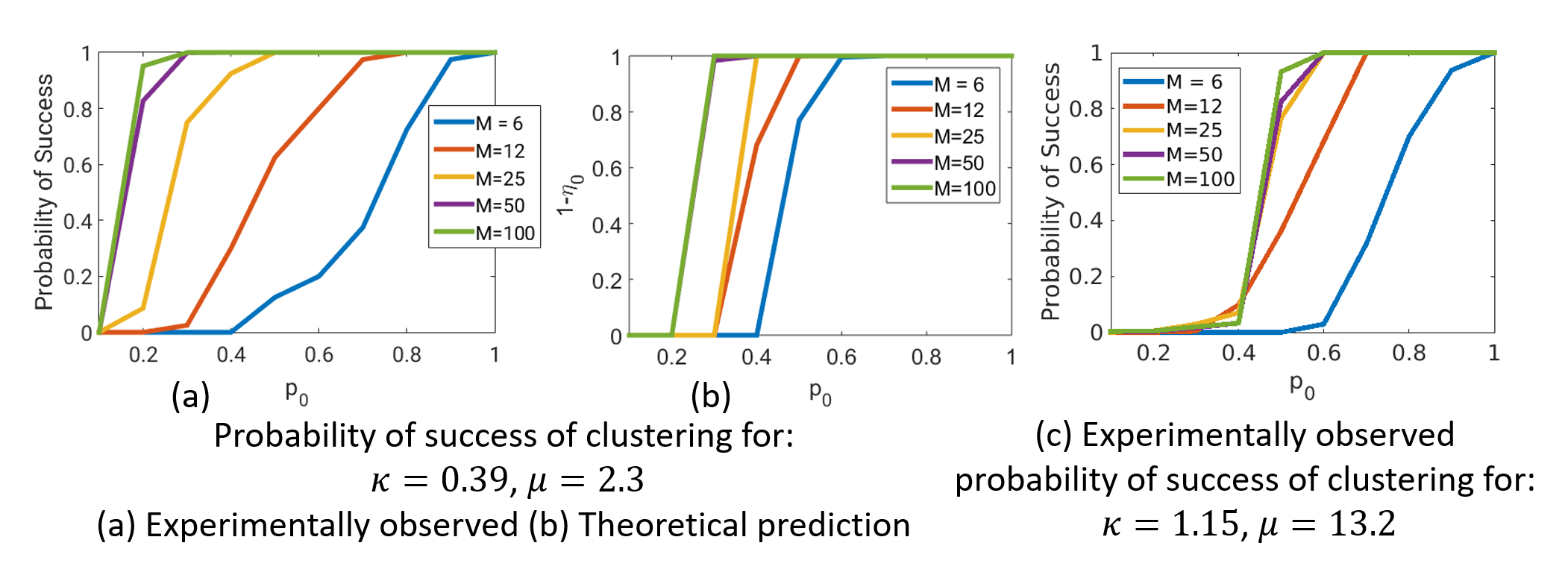}}
	\caption{Experimental results for probability of success. Guarantees are shown for a simulated dataset with $K=2$ clusters. For (a) and (b), $\kappa = 0.39$ and $\mu_0 = 2.3$. (a) and (b) show the experimental and theoretical values  for the probability of success respectively. (c) shows the experimentally obtained probability of success for a more challenging dataset with $\kappa = 1.15$ and $\mu_0 = 13.2$. We do not have theoretical guarantees for this case, since our analysis assumes $\kappa < 1$.}
	\label{FigExpPlots}
\end{figure}

We simulated datasets with $K=2$ disjoint clusters in $\mathbb R^{50}$ with a varying number of points per cluster. The points in each cluster follow a uniform random distribution. We study the probability of success of the $H_1$ penalty based clustering algorithm as a function of $\kappa$, $M$ and $p_0$. For a particular set of parameters the experiment was conducted $20$ times. Fig \ref{FigExpPlots} (a) shows the result for datasets with $\kappa = 0.39$ and $\mu_0 = 2.3$. The theoretical guarantees for successfully clustering the dataset are shown in (b). Our theoretical guarantees hold for $\kappa < 1$. However, we demonstrate in (c) that even with $\kappa = 1.15$ and $\mu_0 = 13.2$, our clustering algorithm is successful.

\begin{figure}
	\centering
	\center{\includegraphics[width=0.48\textwidth]{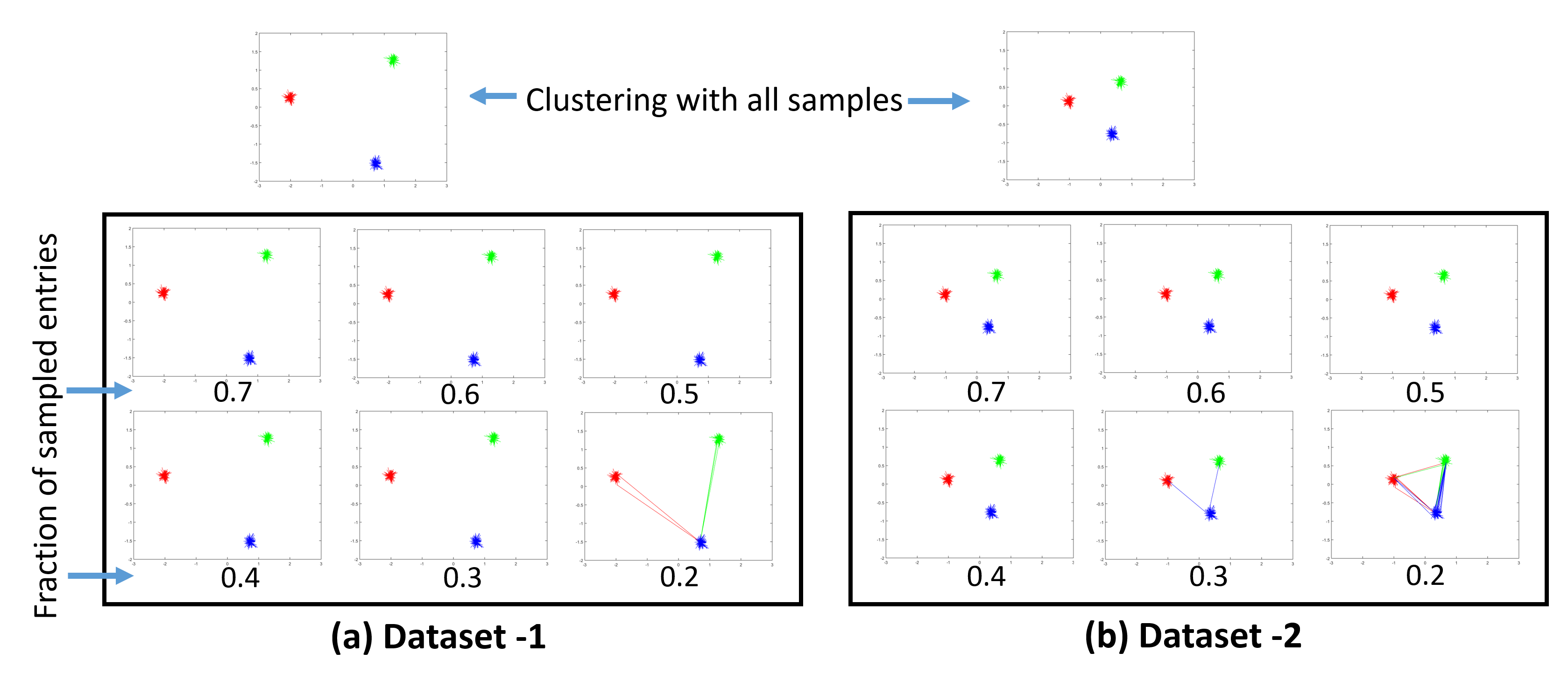}}
	\caption{Clustering results in simulated datasets. The $H_1$ penalty is used to cluster two datasets with varying fractions of missing entries. We show here the 2 most significant principal components of the solutions. The original points $\{\mathbf x_i\}$ are connected to their cluster centre estimates $\{\mathbf u_i^*\}$ by lines.}
	\label{FigSim1}
\end{figure}

Clustering results with $K=3$ simulated clusters are shown in Fig \ref{FigSim1}. We simulated Dataset-1 with $K=3$ disjoint clusters in $\mathbb R^{50}$ and $M=200$ points in each cluster. For each of these $3$ cluster centres, $200$ noisy instances were generated by adding zero-mean white Gaussian noise of variance $0.1$. The dataset was sub-sampled with varying fractions of missing entries ($p_0=1,0.9,0.8,\ldots,0.3,0.2$). We also generate Dataset-2 by  halving the distance between the cluster centres, while keeping the intra-cluster variance fixed. We test the proposed algorithm on these datasets using the $H_1$ penalty. Since the points lie in $\mathbb R^{50}$, we take a PCA of the points and their estimated centres and plot the $2$ most significant components. The $3$ colours distinguish the points according to their ground-truth clusters. Each point $\mathbf x_i$ is joined to its centre estimate $\mathbf u_i^*$ by a line. We observe that the clustering algorithm is more stable with fewer missing entries. 

\subsection{Clustering of Wine Dataset}

\begin{figure}[!t]
	\centering
	\center{\includegraphics[width=0.3\textwidth]{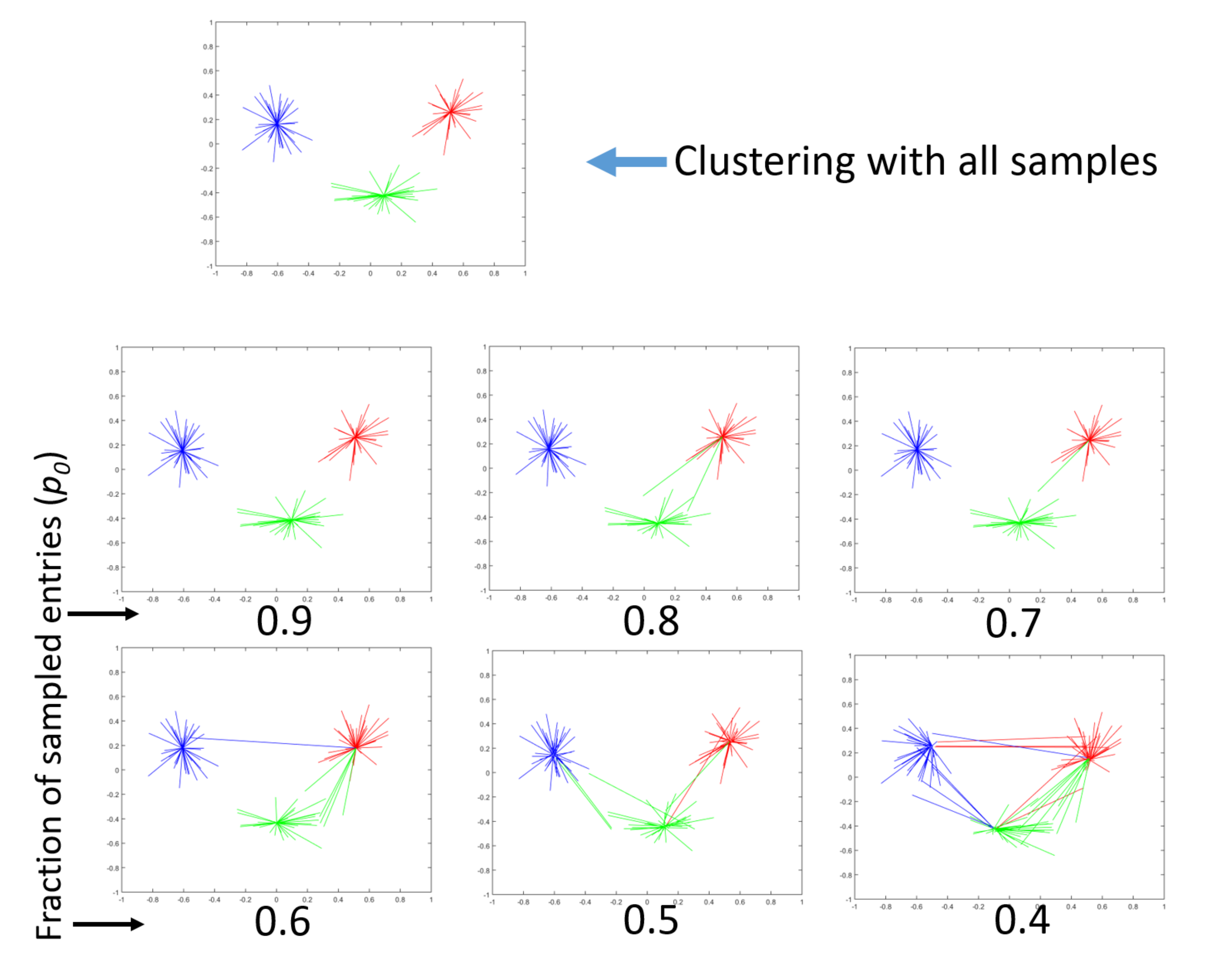}}
	\caption{Clustering the Wine dataset. The $H_1$ penalty is used for clustering with varying fractions of missing entries.}
	\label{FigWine1}
\end{figure}

We apply the clustering algorithm to the Wine dataset \cite{Lichman:2013}. Each data point has $P=13$ features. We created a dataset without outliers by retaining only $M=40$ points per cluster, resulting in $120$ points. The results are displayed in Fig \ref{FigWine1} using the PCA technique as explained in the previous sub-section. It is seen that the clustering is quite stable and degrades gradually with increasing fractions of missing entries.

\vspace{-1.5em}
\section{Conclusion}
We propose a clustering technique that can handle the presence of missing feature values. We derive theoretical guarantees for the successful recovery of the clusters using the proposed optimization problem. We also propose an algorithm to efficiently solve a relaxation of the above problem. This algorithm is demonstrated on simulated and real datasets. It is observed that the proposed scheme can perform clustering even in the presence of a large fraction of missing entries.

\vspace{-1.5em}

\bibliographystyle{IEEEtran}
\bibliography{refs}

\end{document}